\pdfoutput=1

\documentclass[11pt]{article}

\usepackage[final]{emnlp2023}

\usepackage{times}
\usepackage{latexsym}
\usepackage{bbm}
\usepackage[T1]{fontenc}
\usepackage[utf8]{inputenc}
\usepackage{microtype}

\usepackage{inconsolata}
\usepackage{hhline}
\usepackage{hyperref}
\usepackage{diagbox}
\usepackage{graphics}
\usepackage{multirow}
\usepackage{algpseudocode}
\usepackage{algorithm}
\usepackage{graphicx}
\usepackage{pdfpages}
\usepackage{makecell}
\usepackage{tabularx}
\usepackage{float}
\usepackage{siunitx}
\usepackage{amsmath}
\usepackage[T1]{fontenc}

\usepackage[utf8]{inputenc}

\usepackage{microtype}
\usepackage{dcolumn}
\newcolumntype{d}[1]{D{.}{.}{#1}}
\usepackage[normalem]{ulem}
\usepackage{booktabs}
\usepackage{siunitx}
\DeclareMathOperator{\T}{T}

\DeclareMathOperator{\D}{\mathcal{D}}

\DeclareMathOperator{\gt}{gt}
\DeclareMathOperator{\PMI}{PMI}
\DeclareMathOperator{\PELM}{PELM}
\DeclareMathOperator{\len}{len}

\newif\ifdraft
\draftfalse
\ifdraft
\newcommand{\wh}[1]{{\color{magenta}#1}}
\newcommand{\jc}[1]{{\color{cyan}#1}}
\newcommand{\kl}[1]{{\color{red}#1}}
\newcommand{\cc}[1]{{\color{blue}#1}}
\newcommand{\ma}[1]{{\color{orange}#1}}
\newcommand{\jcremove}[1]{\textcolor{cyan}{{\sout{#1}}}}
\else
\newcommand{\wh}[1]{{}}
\newcommand{\jc}[1]{{}}
\newcommand{\kl}[1]{{}}
\newcommand{\cc}[1]{{}}
\newcommand{\ma}[1]{{}}
\newcommand{\jcremove}[1]{{}}
\fi

\title{Toward Joint Language Modeling for Speech Units and Text}

\author{Ju-Chieh Chou\textsuperscript{1}\thanks{\ \ Work done during an internship at Meta AI.},\ Chung-Ming Chien\textsuperscript{1},\ Wei-Ning Hsu\textsuperscript{2},\ Karen Livescu\textsuperscript{1}, \\ {\bf Arun Babu\textsuperscript{2},\ Alexis Conneau\textsuperscript{3}\thanks{\ \ Work done while at Meta AI.},\ Alexei Baevski\textsuperscript{4}\footnotemark[2],\ Michael Auli\textsuperscript{2}} \\ \\
\textsuperscript{1}Toyota Technological Institute at Chicago
\textsuperscript{2}Meta AI \textsuperscript{3}OpenAI \textsuperscript{4}Character AI \\ 
\texttt{jcchou@ttic.edu,wnhsu@meta.com}
} 

\begin{document}
\maketitle

\begin{abstract}
Speech and text are two major forms of human language.
The research community has been focusing on mapping speech to text or vice versa for many years. However, in the field of language modeling, very little effort has been made to model them jointly.
\ma{Maybe stress how there has been lots of work on mapping speech to text (language!) or vice versa but less so for joint models of both and that this paper falls in the latter category.}
In light of this, we explore joint language modeling for speech units and text. Specifically, 
\jcremove{which combines speech units derived from self-supervised speech models with text in a joint language model. }
we compare different speech tokenizers to transform continuous speech signals into discrete units and use different methods to construct mixed speech-text data. 
\ma{Paired speech-text data reminds me of speech utterances paired with their transcriptions but that's not everything you are doing. Maybe say "mixed speech and text data"?} 
We introduce automatic metrics to evaluate how well the joint LM mixes speech and text. We also fine-tune the LM on downstream spoken language understanding (SLU) tasks with different modalities (speech or text) and test its performance to assess the model's learning of shared representations.
Our results show that by mixing speech units and text with our proposed mixing techniques, the joint LM improves over a speech-only baseline on SLU tasks and shows zero-shot cross-modal transferability. 
\wh{maybe you want to give a name to the proposed model instead of referring it to LM, which might be confused with general text based language models}
\cc{Maybe we should say this sentence in a different way since "continuation" itself does not sound like a goal.}

\end{abstract}

\section{Introduction}\label{sec:intro}
\begin{figure}[t]
\centering
\includegraphics[width=\columnwidth]{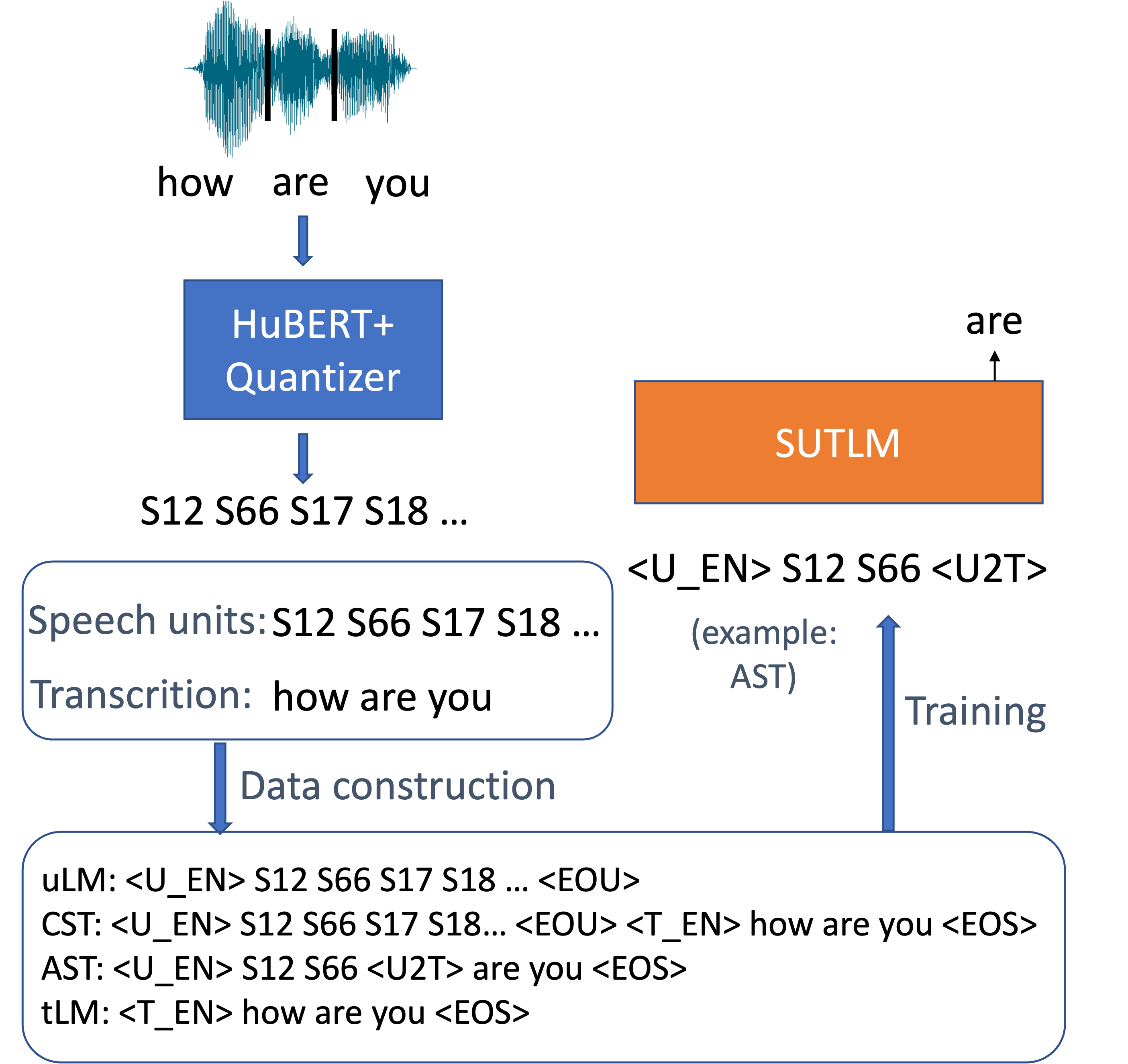}
\centering
\caption{\kl{It might be better to replace the "how are you" at the top with a waveform} An illustration of our workflow. We tokenize speech signals into discrete units and mix them with text to create speech-text data. Our SUTLM is then trained on a combination of speech-only, text-only, and speech-text data. More details on the data formats can be found in~\autoref{tab:format}.
\ma{This figure is nice. I wonder if we should pull it to the first page? It'd be nice to have some figure at the very beginning that illustrates that you build sequence models over sequences consisting both of text and speech units. This could be very schematic and without some of the details you have at the top of the current figures. Anyway, just a suggestion and not required.}
}\label{fig:ast}
\end{figure}
Speech and language processing research has largely focused on spoken and written language separately. 
However, the integration of speech and text in a single model holds potential benefits.
Speech data contains prosodic information that does not exist in the text, which can help in modeling dialogues.
On the other hand, text data from sources like Wikipedia can provide structural knowledge that is not available in most speech datasets. 
\jcremove{, thereby improving the knowledge acquisition ability of models}
Moreover, the amount of written text on the internet exceeds the size of any available speech dataset.

\jc{is this motivation strong enough? } \wh{It is not very strong. We should mention why we think they can help each other. For example, text can be obtained in large quantities than speech; text usually contains more structured knowledge; speech on the other hand capture other types of information better, such as conversation.}

The impressive performance of text large language models (LLMs) has caused a revolution in natural language processing~\cite{radford2019language,brown2020language}. On the other hand, generative spoken language models (GSLM)~\cite{lakhotia2021generative}, which are LMs trained on discrete speech units derived from self-supervised  representations~\cite{hsu2021hubert}, are also promising for spoken language modeling.

In this work, we aim to fill the gap between text-only and speech-only LMs by developing and studying design choices for a joint Speech Unit and Text Language Model (SUTLM). 
For speech, we use a self-supervised learning (SSL) speech model, i.e. HuBERT~\cite{hsu2021hubert}, to convert continuous speech signals into speech units. We then combine the units with text data to train an LM that models speech units and text jointly.
We convert speech-only, mixed speech-text, and text-only data into token sequences (as shown in~\autoref{fig:ast} and~\autoref{tab:format}), and train the model as an LM.
\cc{ADD A PARAGRAPH HERE to connect previous text-only and speech-only LMs.}

\kl{if the backup pars above are not in the paper, then you need to say before this something about how the LM is trained.}\wh{+1}
To evaluate the SUTLM, automatic metrics are developed to quantify the cross-modal ability of the LMs.
We also fine-tune our models on downstream tasks for spoken language understanding. We fine-tune the SUTLMs on either the speech or text data and test them on either speech or text to understand how well the models learn to align the two modalities. 
\jcremove{to evaluate their ability of transferring to downstream tasks.}

Our main contributions are: 
\begin{itemize}
\item We present a
joint autoregressive LM trained on both speech and text (\autoref{sec:method}). 
\item We develop automatic metrics that require no fine-tuning for the evaluation of an SUTLM, and show that the proposed metrics are indicative of the model's cross-modal transfer ability on downstream tasks (\autoref{sec:metric}). \jc{Not sure if we can say that}
\kl{has this been resolved?}\jc{Originally it was predictive or something, I think indicative is okay to use}
\ma{can you strengthen this contribution by saying that the metrics are indicative of downstream performanace or the like?}
\item Empirically, we show that units covering a larger span obtained through SentencePiece tokenization~\cite{kudo2018sentencepiece} outperform local units learned by existing self-supervised models~\cite{hsu2021hubert} (\autoref{sec:unit}).
\item We find that mixing speech units and text with our proposed techniques (\autoref{sec:pair} \& \autoref{sec:ast}) improves the cross-modal ability of the model. \jcremove{to handle both speech and text}(\autoref{sec:eval}). 
\cc{should we also say how do we know the representations are better shared?} \kl{yes, or else describe the contribution differently}
\jcremove{\item Based on results on the SLUE benchmark, mixing speech and text during pre-training improves the model's performance, and also enables zero-shot transfer between speech and text at test time (\autoref{sec:pair} \& \autoref{sec:ast}).}

\end{itemize}


\section{Related Work}

\subsection{SSL speech models}
Self-supervised pre-training enables speech models to learn the information in speech without paired text transcriptions and show impressive performance on tasks such as automatic speech recognition (ASR) with minimal supervised fine-tuning~\cite{baevski2020wav2vec,hsu2021hubert,chen2021wavlm}.
As SSL speech models learn phonetically meaningful speech representations~\cite{pasad2023layer}, they can be used as a feature extractor~\cite{yang2021superb} or a quantizer to transform continuous speech into discrete units~\cite{lakhotia2021generative,lee2021direct,lee2021textless,lin2022dual,chen2022speech}. 
\jcremove{The discretization process disentangles linguistic content from complex paralinguistic information in speech \cite{polyak2021speech}.
The resulted discrete speech units can be used as pseudo-text to improve the performance of speech tasks, such as speech-to-speech translation~\cite{lee2021direct,lee2021textless,chen2022speech} or spoken question-answering~\cite{lin2022dual}, in combination with text data~\cite{lee2021direct,chen2022speech} or even under the textless setting~\cite{lee2021textless}.}
In this work, we use the HuBERT model~\cite{hsu2021hubert} along with a quantizer to tokenize continuous speech into discrete representations. 
The discrete speech units are then combined with text data to train a single LM that is able to model speech and text jointly.

\subsection{Textless NLP}\label{sec:textless}
Textless NLP~\cite{lakhotia2021generative,polyak2021speech,kharitonov2021text} is a framework to model speech in the absence of textual data.
It consists of three components: a speech-to-unit tokenizer, a unit LM (uLM), and a unit-to-speech detokenizer.
The tokenizer takes speech signals as inputs to generate discrete speech units.
A uLM is trained to predict the next token in an utterance given its prior context.
Once the uLM is trained, it can be used to generate unit sequences autoregressively.
In the end, the detokenizer is used to convert the generated unit sequences to speech signals.


\subsection{Joint speech-text transformers}\jc{This section is really long. }

 \kl{it would be good to define encoder-only, encoder-decoder, and decoder-only models, and say a bit more about what are the uses/motivations for each}
Transformer models have been extremely successful in natural language and speech processing~\cite{vaswani2017attention,gulati2020conformer}, with three major configurations: encoder-decoder models~\cite{vaswani2017attention}, encoder-only models~\cite{devlin2018bert}, and decoder-only models~\cite{radford2018improving}.
\jcremove{In general, transformer encoders have access to future context, while decoders follow a left-to-right manner.
As a result, transformer encoders are mostly used as feature extractors to extract representations, and the decoders are mostly used in sequence generation tasks.} 

Previous works on speech-text joint transformers mostly adapt the encoder-decoder~\cite{ao2021speecht5,tang2022unified,cheng2022mu} or encoder-only~\cite{chung2020splat,bapna2021slam,chen2022maestro,zhang2022speechlm} architectures.
Compared with decoder-only architectures, the training of these models typically requires multiple losses and explicit alignments between paired speech and transcriptions.
This makes the hyper-parameter selection time-consuming.
Also, encoder-only and encoder-decoder models are mostly used in the pre-training + fine-tuning paradigm, which limits the use cases of these models.

On the other hand, decoder-only models on text~\cite{radford2019language,brown2020language} show the impressive capability of in-context learning, which also reduces the efforts spent on fine-tuning pre-trained models.
In light of this,
we explore decoder-only models for speech-text joint training.
\kl{why has the VALL-E bit been removed?} 
In this under-explored area, the concurrent work VALL-E~\cite{wang2023neural} is the only other attempt to build a decoder-only model jointly modeling speech and text.
However, VALL-E's purpose is controllable text-to-speech synthesis (TTS), and the work
mainly focuses on the acoustic controllability of the generated speech
, while our work aims to build a general-purpose joint LM and mainly focuses on modeling the content of spoken language.




\section{Method}
\label{sec:method}
We start with a dataset of sentences $\D=\{s^1, s^2, \dots, s^n\}$, where a sentence $s^i$ is composed of a sequence of $T_i$ tokens $(z_1^i, z_2^i, \dots, z_{T_i}^i)$, where $z_j^i$ can be either text or speech units.
The SUTLM is trained to predict the next token $z^i_j$  
\jc{change the order of sentences} given its prior context $z^i_{<j}$.
We maximize the log-probability of the data
\begin{equation}
\label{eq:loss}
\sum_{i=1}^n \sum_{j=1}^{T_i} \log P(z^i_j|z^i_{<j})
\end{equation}
\jcremove{The tokens $z_j^i$ can be referred to as either speech or text units.}
In the following sections, we describe how we construct token sequences from speech and text.
\jcremove{In section~\autoref{ssec:uLM} and~\autoref{ssec:tLM} we use unpaired data, and in section~\autoref{ssec:CST} and~\autoref{ssec:AST} we talk about constructing token sequences with paired speech and text.}
An example of our data formats can be found in~\autoref{tab:format}.


\begin{table*}[h!]
\begin{tabular}{l|l}
Task & Example                                                                                                                                                 \\ \hhline{==}

uLM  & \verb|<U_EN> S12 S66 S17 S18 ... <EOU> |                                                                              \\
CST  & \verb|<U_EN> S12 S66 S17 S18 ... <EOU> <T_EN> how are you <EOS>|  \\
CST  & \verb|<T_EN> how are you <EOS> <U_EN> S12 S66 S17 S18 ...<EOU>| \\
AST  & \verb|<U_EN> S12 S66 <U2T> are you <EOS>|                                          \\
AST  & \verb|<T_EN> how <T2U> S17 S18 ... <EOU>|                                          \\
tLM  & \verb|<T_EN> how are you <EOS> |                                                                            
\end{tabular}
\caption{An example of the formats of unpaired (uLM, tLM) and mixed speech-text (CST, AST) data. For the CST and AST formats, speech units and text can be present in a sequence in different orders . \texttt{<U\_EN>} and \texttt{<T\_EN>} are used at the beginning of the unit/text sequence. \texttt{<EOU>} and \texttt{<EOS>} are used at the end of the unit/text sequences. \kl{why "\_EN"?  and why <EOS> for end of text (what is the S for)? } \texttt{<U2T>} and \texttt{<T2U>} are used when switching from unit to text and text to unit at word boundaries. \wh{shouldn't the first task (uLM) ends with EOU? \cc{use a sample that can show how the text tokenizer works?}\jc{prefer not to, this is just an example. } \cc{remove the second sentence since there's no S-AST anymore}}
}
\label{tab:format}
\end{table*}
\subsection{Speech-only: unit LM (uLM)}
\label{ssec:uLM}
Prior work has shown that discrete speech units derived from a pre-trained HuBERT model can be used as compact representations to encode speech content, enabling the training of a unit language model~\cite{lakhotia2021generative}.
However, when combining speech with text, the time scales of speech units and text differ. HuBERT units are typically on the phone or sub-phone level, as shown in \autoref{tab:ratio}\wh{TODO: add ref}. This leads to longer sequences, making it difficult for the model to capture long-term dependencies.
On the other hand, subword tokenizers for text generally break text sequences into chunks of a larger size than speech units.
This length mismatch between speech and text makes it challenging to model them in a single model.
Therefore, we use a subword tokenizer~\cite{kudo2018sentencepiece} to combine HuBERT units into larger chunks as in~\cite{wu2022wav2seq} to mitigate the length mismatch.
\jcremove{We also expect the subword tokenizer to capture the common patterns in speech unit sequences.}

The process of generating speech units is as follows. 
Speech signals are first fed into a HuBERT model.\wh{TODO: add we deduplicate the units} 
The representations in the final layer 
are then clustered with the k-means algorithm. The cluster IDs are used as the discrete speech units after removing consecutive repeating units~\cite{lakhotia2021generative}.\footnote{For example, the unit sequence \texttt{13 13 15 80 80 80} becomes \texttt{13 15 80} after removing repetitions.}
These units are then further combined by the subword SentencePiece tokenizer~\cite{kudo2018sentencepiece}. 
\jcremove{Here, we try several different vocabulary sizes with the SentencePiece tokenizer.} The resulting average number of tokens per second can be found in~\autoref{tab:ratio}.
\jcremove{When using a vocaulary size of 10k, the sequence length of the combined speech units can be shortened by 47\%, as shown in~\autoref{tab:ratio}, reducing the length mismatch between text and speech units.}

\begin{table}[]
\centering
\begin{tabular}{l|c}
        \hline
       & \multicolumn{1}{l}{Average tokens per second} \\ \hhline{==}
Phone & 20.32 \\\hline
HuBERT & 50.00     \\
\ \  + deduplication & 33.33                              \\
\ \ \ \  + SP 10k & 17.67                             \\
\ \ \ \  + SP 32k & 14.33                           \\
\hline
\end{tabular}
\caption{The average number of tokens per second for different types of speech units. SP 10k and 32k refer to SentencePiece tokenization~\cite{kudo2018sentencepiece} applied to HuBERT units to create a dictionary with 10k and 32k tokens respectively.\jcremove{, and use these tokens as the speech units.}}\wh{TODO: provide an average frame rate. include a row of non-deduplicated HuBERT units (50Hz)}
\label{tab:ratio}
\end{table}

\subsection{Text-only: text LM (tLM)}
\label{ssec:tLM}
We train another SentencePiece tokenizer~\cite{kudo2018sentencepiece} using the text-only corpus~\autoref{sec:textonly} to convert text into subword tokens. 
The resulting vocabulary size of the subword tokens is around 45k. 

\subsection{Concatenated speech-text (CST)}
\label{ssec:CST}
To present paired speech-text data to the SUTLM, we first convert speech units and their transcriptions into the uLM and tLM formats, respectively, and combine them into one sequence by simply concatenating them as shown in~\autoref{tab:format}.
\jcremove{ASR concatenates the text transcription of an utterance after its corresponding speech units.  Similarly, TTS concatenates the speech units of an utterance after its corresponding transcription.}
The CST format explicitly tells the model 
the correspondence between paired speech and text and thus encourages the model to learn the dependence between speech units and the corresponding text transcriptions.
\jcremove{A CST sequence is built in the following manner.Given an utterance and its paired transcription, we first convert them into a speech unit sequence and a text token sequence, respectively, with the uLM and tLM method described in the previous sections.
The two sequences are then concatenated to form the CST sequence.}
\jcremove{For 50\% of the time, we have speech units before text to form the ASR data; otherwise we have text before speech units, which results in the TTS data format.
}

\subsection{Alternating speech-text (AST)}
\label{ssec:AST}
Aside from simply concatenating the sequences of speech units and text, we also construct mixed speech-text that takes the word-level correspondence into consideration.

We use a pre-trained speech recognizer~\cite{mcauliffe2017montreal} to force-align speech and its transcription to obtain the word boundaries in an utterance.
We then randomly sample some word boundaries within the utterance\footnote{For a sentence with $k$ words, we uniformly sample $\lfloor N \rfloor$ boundaries as the switching points with $N\sim\mathcal{N}(\frac{k}{10}, 1)$. \kl{I guess there is some rounding involved?}\jc{yes, added.}} as the "switching points", which divide the utterance into several chunks.
The alternating speech-text (AST) sequence is then constructed by alternatively filling in the chunks with uLM speech units and tLM text tokens, resulting in a sequence that switches modalities at every switching point. 
Special tokens \texttt{<U2T>} and \texttt{<T2U>} are inserted when switching from speech units to text and text to speech units, respectively. 
\jcremove{AST data encourage the model to learn shared representations because it has to output the same distribution over tokens given either speech or text corresponding to the same content.}
\jcremove{For example, as shown in ~\autoref{fig:ast}, the model has to predict "\texttt{are}" after either "\texttt{S12 S66}" or "\texttt{how}", so it has to encode "\texttt{S12 S66}" and "\texttt{how}" into similar representations.
The AST data thus provide stronger guidance to the language model to relate speech units to text.}
\kl{and why do you think ASR or TTS should work?}\jc{telling the model what is the corresponding speech/text directly}




\section{Evaluation Metrics}\label{sec:metric}

\wh{As for fine-tuning based versus fine-tuning free metrics, you can argue that while fine-tuning is a typical way to evaluate how well a model is pre-trained by citing w2v2, hubert, wavlm, etc, it takes much longer to iterate and might also be less reliable because the hyperparamter might favor one pre-trained model over the other if extensive fine-tuning hyperparameter sweeping is not performed.}
We introduce automatic metrics that require no fine-tuning to evaluate the SUTLM. 
Fine-tuning is a common approach to assess the quality of pre-trained models~\cite{baevski2020wav2vec,hsu2021hubert,chen2021wavlm}.
However, it is a time-consuming process and the reliability of the experiments highly depends on the hyper-parameter selection process. Furthermore, there is no reliable metric to measure the cross-modal ability of LMs. 
\jcremove{Previous work also evaluates the content of generated speech with an external LM~\cite{lakhotia2021generative}.}
\jcremove{Such model-based metrics eliminate the uncertainty of fine-tuning but still takes extra efforts to build the external LM.}
In light of this, we propose Context Retrieval Accuracy (CRA), a new  metric that does not require fine-tuning, to evaluate the cross-modal ability of an SUTLM.
\jcremove{, as a strong support to the model-based and fine-tuning-based evaluations.}


\wh{I think it would be better to first state ``what would be considered a good model'' and then propose ``which metrics to use''. For example, you can argue that a good LM model learns the sentence structure better, the conditional probability of a sentence should be higher given a proper previous sentence compared to given a random sentence. Hence, this motivate the design of CRA. Furthermore, you can argue that we want the proposed model to produce similar representation for the same content regardless of the modality; hence we consider cross-modal CRA where context and the continuation are in different modalities.}

\subsection{Context Retrieval Accuracy (CRA)\kl{I wonder if there's a more intuitive name for this}}\jc{contrastive matching accuracy?\wh{what about ``Context Retrieval Accuracy''}}\label{sec:cra}
The motivation of Context Retrieval Accuracy (CRA) comes from the intuition that a good LM should learn to predict the next token based on its prior context.
When we divide a sentence into prompt and continuation, a good LM should be able to capture the dependence between them. That is, it should assign a higher conditional probability to the continuation given its corresponding prompt than given a random prompt.

To measure CRA, we gather a collection of $m$ sentences $\mathcal{C} = \{s^1, s^2, \dots, s^m\}$ and break $s^i$ into a pair of prompt $x^i$ and continuation $y^i$.
Given an SUTLM parameterized by $\theta$, we can measure the conditional probabilities $P_{\theta}(y^i|x^i)$ with \autoref{eq:loss}. The CRA is then computed as:
\begin{equation}
    \frac{1}{m}\sum_{i=1}^m \mathbbm{1}[\arg \max_{j \in \{1 \dots m\}} P_{\theta}(y^i |x^j) = i],
\end{equation}
\jcremove{where $\mathcal{J}$ is a set of indices including $i$ and some randomly sampled indices.}
That is, the LM is used as a scorer to classify whether the matched prompt-continuation pair has the highest conditional probability among a pool of unmatched prompts.

 \wh{TODO: anothe motivation - to compare across different units}

 \kl{it might help to describe this directly as the accuracy of a classifier of prompt given continuation.  also need to say at some point that the prompts have equal length, and directly say why you are not doing classification of continuation given prompt.} \jc{For the classifier point of view, we need to assume uniform P(x)}

\jcremove{We define the CRA as the accuracy of identifying the correct prompt from the pool instead of identifying the correct continuation because}
CRA also has a pointwise mutual information (PMI) interpretation:
\begin{equation}
\begin{split}
    &\arg \max_{j \in \{1 \dots m\}} P_{\theta}(y^i|x^j) = i \\
\implies &\log P_{\theta}(y^i|x^i) \geq \max_{j \in \{1 \dots m\}}\log P_{\theta}(y^i|x^j) \\    
\implies &\log \frac{P_{\theta}(y^i|x^i)}{P_{\theta}(y^i)} \geq \max_{j \in \{1 \dots m\}} \log \frac{P_{\theta}(y^i|x^j)}{P_{\theta}(y^i)} \\
\implies &\PMI(x^i, y^i) \geq \max_{j \in \{1 \dots m\}}\PMI(x^j, y^i) \\
\end{split}
\end{equation}
That is, correctly identifying the prompt implies the matched prompt-continuation pair has a higher PMI than all unmatched prompt-continuation pairs\jcremove{in the pool according the joint probability defined by the SUTLM.}. 
\wh{TODO: mention the prompt/cont can be either modality to evalaute intra/inter-modal continuation}\jc{done}

Ideally, the model should produce similar representations given the same content regardless of the modality. 
Hence, in addition to the uni-modal CRA, we also consider cross-modal CRA, where the prompt and the continuation are in different modalities.
In practice, for example, when we use text as the prompts and speech units as the continuations, we set the probability of emitting text tokens to zero and re-normalize the probability to ensure that the continuation $y^i$ can be only speech units.
Cross-modal CRA can be used as a way to measure whether the SUTLM successfully learns shared representations between text and speech.\jcremove{through mixing speech and text.}

\subsection{Perplexity under External LM (PELM)}
Following previous work, we use the perplexity under external LM (PELM) to measure the quality of the content of generated samples~\cite{lakhotia2021generative}. 
We sample a continuation from the SUTLM given each ground truth prompt.
We then use an external text LM, OPT-6.7B~\cite{zhang2022opt}, to compute the perplexity of the sequence:
\begin{equation}\label{eq:pelm}
\begin{split}
\hat{y}^i &\sim P_{\theta}(y | x^i)\\
x^{\prime i}, y^{\prime i} &=\T(x^i \mathbin\Vert \hat{y}^i) \\
\PELM(\theta)&=2^{\dfrac{-\sum_{i} \log P_{\text{OPT}}(y^{\prime i}| \gt(x^{i}))}{\sum_i \len( y^{\prime i})}}
\end{split}
\end{equation}
where $x^i$ and $\hat{y}^i$ refer to the prompt and sampled continuation, and $\theta$ are the parameters of the SUTLM.
Similarly to cross-modal CRA, we control the modality of sampled continuations by zeroing out the probability of the tokens in the undesired modality.
Since the prompt and the continuation can be either speech units or subword text tokens, we use a transcriber $\T(\cdot)$ to transcribe the concatenated sequences $x^i \mathbin\Vert \hat{y}^i$ into text $x^{\prime i}, y^{\prime i}$.\footnote{For both speech units and text tokens, we first invert the SentencePiece tokenization process to get raw HuBERT units and raw text. For speech units, we further use a 12-layer Transformer encoder with a CTC head to map HuBERT units to text. The transformer is trained on LibriSpeech, with a WER of 5.18\% on dev-clean, and 11.61\% on dev-other.}
$\gt(\cdot)$ is a function that outputs a ground truth transcription when the input is speech units and is an identity function when the input is text.
The external LM is then used to measure the perplexity of the continuation part of the text sequence.

\wh{TODO: Eq 3 is not accurate when x are speech units or when our LM and the external LM uses a different tokenizer.}

\subsection{Evaluation on SLUE tasks} 

We use the SLUE benchmark~\cite{shon2022slue} to evaluate our models on downstream tasks. \jcremove{compare the performance of the learned representations on language understanding tasks, and validate their ability to learn representations that are transferable between speech and text.}
The benchmark includes two tasks, sentiment analysis (SLUE-SA) and named entity recognition (SLUE-NER), with both speech data and transcriptions provided.
After pre-training the SUTLM, we fine-tune it on the SLUE dataset with either speech or text data as inputs to predict the ground-truth labels, and then evaluate it on either speech or text inputs. We evaluate the model on different input modalities to understand the cross-modal ability of the model as in~\cite{hsu2022u,bapna2021slam,bapna2022mslam}. Fine-tuning details can be found in~\ref{sec:downstream}.
\kl{not clear what it means to "fine-tune our LM on SLUE" -- need to say more about the model for each task}
\jcremove{A model which learns shared representations should be able to transfer the supervisory signals of the fine-tuning task from one modality to the other one~\cite{hsu2022u,bapna2021slam,bapna2022mslam}. }

\section{Experiments}
\subsection{Data}
\kl{I don't think you need to say why you don't use audiobooks, unless it is a contribution of the paper that you are not using audiobooks} 
\jc{is it convincing?} \kl{no :)  what are you trying to convince people of?}\jc{the reason to subsample. }

\subsubsection{Speech-only}
We use 5\% of the dataset used in~\cite{aghajanyan2023scaling} to match the size of the mixed speech-text and text-only data. The dataset includes Multilingual LibriSpeech (MLS)~\cite{pratap2020mls}, VoxPopuli~\cite{wang2021voxpopuli}, CommonVoice~\cite{ardila2019common} and Spotify Podcast \& People’s Speech~\cite{aghajanyan2023scaling}.
The subsampled dataset consists of 65k hours of speech.

\subsubsection{Mixed speech-text (CST and AST)}
We use MLS~\cite{pratap2020mls} and VoxPopuli~\cite{wang2021voxpopuli} to create mixed speech-text data without subsampling. The dataset contains 45k hours of speech and 2.7B of words.  
\jc{say something about: the impact of adding text to speech remains unknown}
\subsubsection{Text-only}\label{sec:textonly}
We combine OPT web data~\cite{zhang2022opt}, Wikipedia, and LibriLM~\cite{panayotov2015librispeech}, and then subsample 5\% of it, resulting in a total of 8.5B subwords. 

\subsection{SSL speech tokenizer}\label{sec:speech_unit}
We use a HuBERT Base model trained on 221K hours of unlabeled speech in 8 languages as in~\cite{hsu2022revise, nguyen2023expresso}.\footnote{\url{https://dl.fbaipublicfiles.com/hubert/mhubert_base_vp_mls_cv_8lang_it3.pt}}
After pre-training, the representations at the last layer (12th) are clustered with k-means using 2000 clusters.\jcremove{with duplicated cluster centroids removed. }

\subsection{Model architecture and training}\label{sec:arc}
We use the 24-layer transformer implementation in fairseq~\cite{ott2019fairseq} with 16 attention heads.
The embedding size is 1024, the feed-forward dimension is 4096, and the dropout probability is set to 0.1.
The weights of the embedding layer are tied to the output layer~\cite{press2016using}. The model contains 350M parameters. 

The model is trained for 500k updates on 32 V100 GPUs with a batch size of 8192 tokens per GPU.
We use Adam optimizer~\cite{kingma2014adam} with ($\beta_1, \beta_2$) = (0.9, 0.95).
Gradient clipping with a threshold 1.0 and weight decay of 0.1 are applied to stabilize the training. 
Since the data size is different for different data formats, we resample speech-only, speech-text, and text-only data equally (1/3 for each in every training batch) to prevent the model from being biased toward any of them. 

\subsection{Evaluation setup}\label{sec:eval}
\subsubsection{Automatic Metrics}
We use a subset of the Multilingual LibriSpeech~\cite{pratap2020mls} dev set to evaluate the SUTLM.
To provide enough context to the SUTLM, we filter out sentences of less than 20 words.
For each sentence, we use the first 10 words as the prompt and the rest as continuation.
For the CRA experiments, we evaluate the SUTLM with the 100 shortest utterances in the filtered dataset, while for the PELM experiments, we use the 500 shortest utterances. We use fewer utterances in CRA experiments as the computation of CRA is $O(N^2)$ for $N$ utterances.
We constrain ourselves to sentences with moderate lengths because the continuation part becomes less coherent with the prompt as the sequence length grows, which hurts the sensitivity of the proposed metrics.

When sampling the speech or text continuations in the PELM experiments, we use temperature $t=0.6$ and nucleus sampling~\cite{holtzman2019curious} with $p=0.95$, and truncate the continuation to 10 words (identical to the length of the prompts).
\subsubsection{Downstream Tasks}\label{sec:downstream}
For SLUE-SA, we fine-tune SUTLM by adding a self-attention pooling layer on top of the transformer model after removing the last output layer~\cite{shon2022slue}.
We fine-tune it with a learning rate of 3e-5 for 30k updates and evaluate it with Macro F1~\cite{shon2022slue}.

For SLUE-NER, we follow the SLUE official baseline to formulate the task as an ASR problem and train our model to decode special tokens around each named entity~\cite{shon2022slue}.
We concatenate the output (the text transcription with special tokens before and after each named entity) after the input (speech units when fine-tuned on speech, text tokens when fine-tuned on text) and fine-tune our SUTLM as an LM with the same loss function as \autoref{eq:loss}.
The loss is only applied to the output part of the sequence.
We fine-tune the SUTLM with a learning rate of 3e-5 for 50k updates.
During decoding, we use a beam size of 5 to generate the outputs and evaluate them with Micro F1~\cite{shon2022slue}.
For both SLUE tasks, we report results on the dev set since the test set is not publicly available. We use the fine-tuned HuBERT as the baseline as in~\cite{shon2022slue}.

\subsection{Results}
\begin{table*}[h!]
\centering
\resizebox{2.1\columnwidth}{!}{
\begin{tabular}{c|c|c|c|c|c|cc|cc|cc|cc}
\multicolumn{6}{l|}{} & \multicolumn{2}{c|}{u2u}                           & \multicolumn{2}{c|}{t2u}       & \multicolumn{2}{c|}{u2t}       & \multicolumn{2}{c}{t2t}        \\ \hhline{==============}
row & unit & uLM & CST & AST & tLM           &  \multicolumn{1}{r}{CRA} & \multicolumn{1}{r|}{PELM}  & CRA & \multicolumn{1}{r|}{PELM}   & CRA & \multicolumn{1}{r|}{PELM}  &  CRA & \multicolumn{1}{r}{PELM} 
\\ \hhline{==============}
\multicolumn{6}{c|}{Ground truth continuation} &-&-&-&-&-&-&-&101.4  \\ \hhline{==============}
\textbf{(A)} & HuBERT & v &&& & 1.00 & 193.3 &-&-&-&-&-&- \\ \hline
\textbf{(B)} & SP 10k & v&&& & 0.96 & 163.6 &-&-&-&-&-&- \\ \hline
\textbf{(C)} & SP 32k & v &&& & 0.96 & 177.4 &-&-&-&-&-&- \\ \hhline{==============}
\textbf{(D)}& SP 10k & v &&& v & 0.94 &175.9 & 0.03 & 394.9 & 0.01 & 1973.3 & 0.20$^{\ast\ast}$ & 20.7$^{\ast\ast}$ \\ \hline
\textbf{(E)} & SP 10k & v &  v & &                           & 0.95 & 166.0 &                                   0.37 & 39.1$^{\ast}$ &  0.26 &        43.4$^{\ast}$        & 0.56& 34.7 \\  \hhline{==============}
\textbf{(F)} & SP 10k& v & v & v & v       & 0.97 &  162.8  & 0.70 &  124.7 &  0.81 &  38.7 &  0.67 &  28.2                  
\end{tabular}
}
\caption{
Automatic metrics (CRA and PELM). "u2t" denotes that the prompts are speech units and the continuations are text, and so on. (*): for cross-modal cases (u2t and t2u) in row \textbf{(E)}, the PELM is low because the continuation simply repeats the prompt. We discuss this issue in~\autoref{sec:limit}. (**): The low CRA for t2t is due to the use of MLS as an evaluation set, resulting in a distribution mismatch from the text-only training data. Similarly, the use of OPT data to train the SUTLM results in better PELM on t2t in row (D).
\cc{the HuBERT here refers to deduplicated HuBERT}
}\label{tab:ppl_pair}
\end{table*}
\begin{table}[h!]
\centering
\resizebox{\columnwidth}{!}{
\begin{tabular}{c|c|c|cc|cc}
& & FT data     & \multicolumn{2}{c|}{SP}                           & \multicolumn{2}{c}{TXT}                                                 \\ \hhline{=======}
row & unit &  Eval set    & \multicolumn{1}{l}{SP} & \multicolumn{1}{l|}{TXT} & \multicolumn{1}{l}{SP} & \multicolumn{1}{l}{TXT} \\ \hhline{=======} 
\multicolumn{3}{c|}{Baseline} & 0.46 &-&-&- \\ \hhline{=======}
\textbf{(A)} & HuBERT & uLM     & 0.51  & - & - & -   \\ \hline
\textbf{(B)} & SP 10k & uLM     & 0.56   & - & - & -   \\ \hline
\textbf{(C)} & SP 32k & uLM     & 0.54   & - & - & -   \\ \hhline{=======}
\textbf{(D)} & SP 10k & \makecell{uLM+tLM} &           0.52         &        0.33              &        0.35            &    0.49                   \\ \hline
\textbf{(E)} & SP 10k & uLM+CST     & 0.48                   & 0.42                     & 0.51                   & 0.52                                        \\ \hline
\hhline{=======}
\textbf{(F)} & SP 10k & \makecell{uLM+CST\\+AST+tLM}   & 0.49  & 0.43                   &          0.52            &  0.56                                                           \\

\end{tabular}
}
\caption{Macro F1 score on SLUE-SA. FT data indicates the model is fine-tuned on speech (SP) or text (TXT). Eval set denotes the fine-tuned model is tested on speech (SP) or text (TXT).
}\label{tab:pair_sa}
\end{table}
\begin{table}[h!]
\centering
\resizebox{\columnwidth}{!}{
\begin{tabular}{c|c|c|rr|rr}
& & FT data     & \multicolumn{2}{c|}{SP}                           & \multicolumn{2}{c}{TXT}                                             \\ \hhline{=======}
row & unit & Eval set    & \multicolumn{1}{c}{SP} & \multicolumn{1}{c|}{TXT} & \multicolumn{1}{c}{SP} & \multicolumn{1}{c}{TXT} \\  \hline
\multicolumn{3}{c|}{Baseline} & 54.5 &-&-&- \\ \hhline{=======}
\textbf{(A)} & HuBERT & uLM  & 62.9  & - & - & -\\ \hline
\textbf{(B)} & SP 10k & uLM  & 64.4  & - & - & -\\ \hline
\textbf{(C)} & SP 32k & uLM  & 62.5  & - & - & -\\\hhline{=======}
\textbf{(D)} & SP 10k  & \makecell{uLM+tLM} & 63.2                   & 1.5                     & 0.0                   & 66.8 \\ \hline
\textbf{(E)} & SP 10k & uLM+CST     & 65.0                   & 3.6                    & 0.5                   & 79.5                                        \\ \hhline{=======}                            
\textbf{(F)} & SP 10k  & \makecell{uLM+CST\\+AST+tLM}     & 66.6                   & 25.2                    & 0.3                   & 77.2                                        \\ 
\end{tabular}
}
\caption{The F1(\%) score on SLUE-NER. FT data indicates the model is fine-tuned on speech (SP) or text (TXT). Eval set denotes the fine-tuned model is tested on speech (SP) or text (TXT). 
}\label{tab:pair_ner}
\end{table}

\subsubsection{What kind of speech units works the best?}\label{sec:unit}

We utilize HuBERT units described in~\autoref{sec:speech_unit} (2000 units) and apply SentencePiece tokenizers on them. Results can be found in rows \textbf{(A)}, \textbf{(B)}, \textbf{(C)} in~\autoref{tab:ppl_pair} for automatic metrics,~\autoref{tab:pair_sa} for SLUE-SA and~\autoref{tab:pair_ner} for SLUE-NER. 

The model trained with SP 10k has the best performance in terms of PELM, SLUE-SA, and SLUE-NER, but slightly worse CRA than the model using the original HuBERT units. For CRA for the u2u case (unit prompt, unit continuation), we hypothesize that the model uses low-level acoustic information to make predictions as the CRAs are nearly 1.0 for all types of speech units. Also, HuBERT uses overlapping windows for neighboring tokens, so the first token of the continuation contains information about the previous token. 


For the speech continuation (PELM) experiments, the SP 10k-based sequences are shorter than HuBERT unit-based sequences, so the model trained with SP 10k (row \textbf{(B)}) can generate more coherent continuations.\jcremove{according to PELM while increasing the vocabulary further to 32k (row \textbf{(C)}) leads to worse performance. }


\subsubsection{Do we need paired data to learn shared representations?}

In this section, we compare models trained with and without paired data to investigate the usefulness of paired data. We can compare the results in row \textbf{(D)} and \textbf{(F)} in~\autoref{tab:ppl_pair} for automatic metrics,~\autoref{tab:pair_sa} for SLUE-SA and~\autoref{tab:pair_ner} for SLUE-NER. For cross-modal cases (u2t and t2u), in terms of automatic metrics, the model trained with unpaired data alone (row \textbf{(D)}) has almost random CRAs and high PELMs, indicating a lack of cross-modal ability. 

Similarly, for SLUE-SA, the model trained with unpaired data alone (row \textbf{(D)}) shows almost random macro F1 scores for a 3-way classification task when tested on the other modality. For SLUE-NER, the model trained without exposure to paired data (row \textbf{(D)}) performs worse than models trained with paired data (row \textbf{(F)}) when fine-tuned on speech and shows no transferability between modalities. Row \textbf{(D)} also performs worse than its speech unit-only counterpart (row 
\textbf{(B)}, showing that the model trained solely on unpaired data does not demonstrate any cross-modal transfer ability between speech and text. 

\subsubsection{Does concatenated speech-text (CST) help learn shared representations? }\label{sec:pair}

The next question we want to answer is whether CST is helpful in learning shared representations.
Building on the previous findings (rows \textbf{(A)}, \textbf{(B)}, \textbf{(C)}), we utilize SP 10k as our speech unit vocabulary and present the results in row \textbf{(E)} in \autoref{tab:ppl_pair} for automatic metrics, \autoref{tab:pair_sa} for SLUE-SA, and \autoref{tab:pair_ner} for SLUE-NER. 
 The results show that, compared to using unpaired data alone (row \textbf{(D)}), the model trained with CST (row \textbf{(E)}) has higher CRAs for u2t and t2u, which indicates that the model captures the relationship between speech and text better than models trained with unpaired data alone. 
 
 For SLUE-SA, the model pre-trained with CST shows comparable performance when fine-tuned on one modality and evaluated on the other. The performance when fine-tuning on text and testing on speech is even better than directly fine-tuning on speech (0.51 vs. 0.48). The reason is likely to be that text data provides a less noisy supervisory signal compared to using speech units.  The model trained with extra speech-text data (row \textbf{(E)}) performs worse than the model trained with only speech units (row \textbf{(B)}). The reason may be similar to the "curse of multilinguality"~\cite{conneau2019unsupervised}, where sharing the capacity of the model with other languages or modalities hurts performance.


 \wh{TODO: I find the results comparing (1) and (3) in Table 5 a bit weird. Why adding S-AST data degrades the performance so much?}\jc{The only reason I can think of is too much noise}
 
 For SLUE-NER, concatenated speech-text improves performance over the model trained with only speech units (row \textbf{(B)}) when fine-tuned on speech. Unlike SLUE-SA, which is a classification task, here we need to generate the corresponding transcription along with the named entity tags for SLUE-NER. Hence, the model (row \textbf{(E)}) fine-tuned on speech benefits directly from the extra speech-text data. 
We discuss the implications of the fine-tuning results further in~\autoref{sec:imp_downstream}. 


For speech / text continuation, when only using concatenated speech-text data (CST) as our mixed data, there are no special tokens (\texttt{<U2T>, <T2U>}) to trigger modality switching. As shown in~\autoref{tab:example}, in the u2t case the model trained with CST simply transcribes the speech prompt into its transcription on u2t and synthesizes the text prompt into speech units, resulting in low PELMs for u2t and t2u in row \textbf{(D)} due to the repetition. PELM fails to reflect the quality of the continuation accurately. We discuss this limitation further in~\autoref{sec:limit}. \jcremove{This limitation also suggests the necessity of using AST for speech/text continuation tasks, as special tokens are needed to trigger modality switching when using the LM to continue for a cross-modal prompt.}

\subsubsection{Does alternating speech-text (AST) help learn shared representations?}\label{sec:ast}

This section discusses the benefits of alternating speech-text (AST). 
\jcremove{in enabling the model to learn shared representations. }The results are presented in (row \textbf{(F)}) in~\autoref{tab:ppl_pair} for automatic metrics,~\autoref{tab:pair_sa} for SLUE-SA, and~\autoref{tab:pair_ner} for SLUE-NER. 

By comparing the results of CRA for t2u and u2t in row \textbf{(F)} with those in row \textbf{(E)} in~\autoref{tab:ppl_pair}, we observe an improvement in CRA when the data is directly constructed to switch modalities on word boundaries. We can also see that CRA is similar for t2u, u2t, and t2t. It suggests that the model learns to match context regardless of modality.


In row \textbf{(F)}, PELM for t2u is lower than PELM for u2u as the text prompt is less noisy than speech units. PELM for u2t is only marginally worse than t2t. This shows that the LM trained with AST can continue a sentence regardless of the modality. The worse PELM for u2u and t2u than for u2t and t2t could be attributed to the recognition errors within our unit transcriber. \jcremove{An example of this can be found in~\autoref{tab:example}.}

Regarding SLUE-SA, we can observe that AST and tLM further improve the cross-modal transfer performance (trained on the text and evaluated on speech, or vice versa) in row \textbf{(F)}. 

In SLUE-NER, row \textbf{(F)} also shows better performance than row \textbf{(E)} when fine-tuned on speech and evaluated on speech. There is also non-trivial speech-to-text transfer (fine-tuned on speech and evaluated on text) in row \textbf{(F)}, showing that AST helps in learning transferable features between modalities. 

In SLUE-NER, when fine-tuned on text and evaluated on speech, there is no transferability between speech and text. The reason can be attributed to the fine-tuning task becoming almost trivial. In text NER, in our formulation, the input and output are nearly identical. The only difference is the named entity tags. Further discussion of downstream task performance can be found in~\autoref{sec:imp_downstream}. 
\wh{we may explain that the poor performance of text-to-speech transfer on SLUE-NER might arise from the fact that NER when fine-tuned on text has a trivial task (transcribe the sentence) the encourage the model to throw away shared representation and simply learn copying the input}

\subsection{Limitations of PELM}\label{sec:limit}
We use PELM as a metric to measure the quality of continuations. However, although our SUTLM (row \textbf{(F)}) shows the ability to continue after a cross-modal prompt, the resulting continuation is still only locally consistent as shown in~\autoref{tab:example}. This can be attributed to the use of a 350M-parameter model architecture, which is relatively small in the era of LLMs. 

The PELM metric fails to accurately reflect the result in the case of row \textbf{(E)} when the model simply repeats the prompt. It has been a known phenomenon 
 that LMs tend to assign a high probability to repeated tokens~\cite{holtzman2019curious}. 
 
 To quantify repetition, we compute the proportion of bi-grams in continuations that have appeared in the prompt transcription. For row \textbf{(E)}, the proportions are 0.02, 0.53, 0.42, and 0.02 for u2u, u2t, t2u, and t2t, respectively. For row \textbf{(F)}, the proportions are 0.02, 0.03, 0.01, and 0.03. For row \textbf{(E)}, the continuations for u2t and t2u are simply repeating the content of the prompt. 

We can see that the u2t and t2t PELMs are lower than the ground truth PELM. This is because of the use of the temperature of $0.6$ in the softmax layer, which likely hurts diversity and coherence as in~\cite{caccia2018language,lakhotia2021generative}. \jcremove{Also, from the example in~\autoref{tab:example}, we can see that the generated text is dramatic due to the use of audiobooks as training data. We left the use of more natural data as future work. }

\subsection{Implications for SLU Downstream Tasks}\label{sec:imp_downstream}
We show that mixing speech units and text improves the cross-modal ability of the model. In SLUE-SA, the mixed speech-text data enables the model to zero-shot transfer between speech and text. In SLUE-SA, we remove the output layer from the SUTLM and attach a classification head so the model will always output a valid class. 

In SLUE-NER, using mixed speech-text data directly improves the performance. Since this is a sequence generation task, the mixed speech-text data helps the model generate better text. The transfer from speech to text is non-trivial but not vice versa. This finding aligns with the experiments in~\cite{bapna2022mslam}, in which they also find non-trivial transfer from speech to text but not the other way around. However, we note that different fine-tuning strategies can produce different results, as demonstrated in~\cite{liu2021gpt}.


\section{Conclusion}
Our study on joint language modeling for speech units and text involved developing evaluation metrics and fine-tuning the model on speech and text data. We found that using mixed speech-text data improves the model's cross-modal ability and performance on both automatic metrics and downstream tasks. 

Our study sheds light on the benefits of considering both speech and text in building language models. We hope that this research will motivate the research community to further explore the integration of speech and text data for more comprehensive language modeling. 

Future work in this area could involve investigating the optimal balance between speech and text data in model training and exploring ways to handle multi-modal data beyond the speech-text domain.

\jcremove{Specifically, we compare different speech units and different ways to mix speech-text data. 
We first show that combining speech units using subword tokenizer leads to better performance on speech continuation and downstream tasks. Second, using the automatic metric, we find that 
with our proposed techniques i.e., concatenated speech-text (CST) and alternating speech-text (AST), the model learns shared representations that benefit both downstream tasks and speech/text continuation tasks. For downstream tasks, our proposed techniques enable the model to perform zero-shot transfer between speech and text on a sentiment analysis task, improve a NER task, and show a non-trivial performance on speech-to-text transfer on the NER task. }

\section{Limitations}
Our approach involves using a speech tokenizer that can encode phonetic information (HuBERT) and an off-the-shelf speech recognizer to generate word-level alignment. For other, lower-resource languages, these components may be harder to obtain or may not perform as well. 

For our proposed automatic metrics, the complexity of CRA grows at a rate of $O(N^2)$, which can be expensive when evaluated on a larger number of utterances or when scaling up the model size. PELM, on the other hand, also has limitations as stated in~\autoref{sec:limit}. For the empirical results on downstream tasks, we test our SUTLMs on the SLUE benchmark, which has only two tasks. Extending the experiments to more downstream tasks may provide more insights. 

Finally, we only study relatively small SUTLMs (350M parameters). It is unclear how scaling it up would affect the results. 

\kl{reference formatting needs some cleanup}

\bibliography{anthology,custom}
\bibliographystyle{acl_natbib}

\clearpage
\onecolumn
\appendix

\section{Appendix}\label{sec:appendix}
\begin{table}[h]
\centering
\resizebox{0.92\columnwidth}{!}{
\begin{tabularx}{\columnwidth}{c|cX}

\multicolumn{1}{c|}{\textbf{prompt}} & \multicolumn{2}{l}{they need the same aid and they}                                                                                                                                                                          \\ \hhline{===}
\textbf{model}                  & \multicolumn{1}{l|}{\textbf{modalities}} & \multicolumn{1}{c}{\textbf{Continuation}}                                                                                                                                                           \\ \hhline{===}
\multirow{4}{*}{\textbf{(E)}}      & \multicolumn{1}{c|}{u2u}        & soon will increase and reach the grave more gradually than ever they seek their own home and many little people suffer from the hospitality and care of the poor                                           \\ \cline{2-2}\cline{2-3}
                            & \multicolumn{1}{c|}{u2t}        & they need the same aid and they need the same aid                                        \\ \cline{2-3} 
                            & \multicolumn{1}{c|}{t2u}        &  they need they they need the sameling  the need the need the same sense of will that they need for others others to to to to do they will a and and they to have                                                           \\ \cline{2-3} 
                            & \multicolumn{1}{c|}{t2t}        & need the same help they can make and     are able to make the best of them and the best of them
   \\ \hhline{===}
\multirow{4}{*}{\textbf{(F)}}      & \multicolumn{1}{c|}{u2u}        & certainly do need the same money and they do not give themselves out of either field or settlement but they do need it they say and they need it                                           \\ \cline{2-2}\cline{2-3}
                            & \multicolumn{1}{c|}{u2t}        & shall be able to perform it all the same the king of egypt is not free from the persecution of the jews he is free from the persecution of the jews                                        \\ \cline{2-3} 
                            & \multicolumn{1}{c|}{t2u}        & can no longer face it but though these men believe women now to be the principal duties the progress made will actually continue                                                           \\ \cline{2-3} 
                            & \multicolumn{1}{c|}{t2t}        & will be the same as they have been when they were born and they will be the same as they now are if they were not born they would be the same as they now are 
                            
\\ \hhline{===}
\multicolumn{1}{c|}{\textbf{prompt}} & \multicolumn{2}{l}{barrack was just coming through the doorway carrying half}                                                                                                                                                                          \\ \hhline{===}
\textbf{model}                  & \multicolumn{1}{l|}{\textbf{modalities}} & \multicolumn{1}{c}{\textbf{Continuation}}                                                                                                                                                           \\ \hhline{===}
\multirow{4}{*}{\textbf{(E)}}      & \multicolumn{1}{c|}{u2u}        & doesn papa and several small scotch girls who had been sitting by the fire with him he said to the girls on whom he kept his second mother mother is good to me and so i would not let her go away \\ \cline{2-2}\cline{2-3}
                            & \multicolumn{1}{c|}{u2t}        & barrek was just coming through the doorway carrying half a dozen other things                                        \\ \cline{2-3} 
                            & \multicolumn{1}{c|}{t2u}        &  round of the door he was just coming through the doorway carrying half around of the door they stopped at the back of the doorway and half a around of the doorway and half a round of the doorway                                                          \\ \cline{2-3} 
                            & \multicolumn{1}{c|}{t2t}        & a dozen men and a dozen men and women i believe and i think i could have saved the lives of those who loved me-but i don't know how i could have saved them
   \\ \hhline{===}
\multirow{4}{*}{\textbf{(F)}}      & \multicolumn{1}{c|}{u2u}        & dozen boxes when he saw the stick black inside in his room stepping out of his way into the hall a chuckle of joy fell in the drawing room and he seized the boxes and broke down the door                                           \\ \cline{2-2}\cline{2-3}
                            & \multicolumn{1}{c|}{u2t}        & a cup of coffee when gertie came up with a basketful of rice and a handful of water and then she came to the house of mrs smiths and she said to gertie                                       \\ \cline{2-3} 
                            & \multicolumn{1}{c|}{t2u}        & dozen packs and a wrapper and a light sparkling light across the face of jack and a burning gold bullet and a very sharp thumb                                                           \\ \cline{2-3} 
                            & \multicolumn{1}{c|}{t2t}        & a dozen toy guns and a hundred toy guns and a hundred toy guns 
\end{tabularx}
}
\caption{Example for speech and text continuation. Speech continuation has been transcribed by the transcriber. \wh{TODO: the example seems to be looping a lot. Have you tried other temperature and see if that still occurs?}\jc{For temp=1.0, they loop much less but also much less coherent. }}\label{tab:example}
\end{table}

\end{document}